\documentclass[manuscript]{acmart}

\usepackage{url}
\usepackage{kotex}
\usepackage{graphicx}
\usepackage{booktabs} % For formal tables
\usepackage{multirow}
\usepackage{amsmath}
\usepackage{enumitem}
\usepackage{lingmacros}
\usepackage{lingmacros}

\AtBeginDocument{%
  \providecommand\BibTeX{{%
    \normalfont B\kern-0.5em{\scshape i\kern-0.25em b}\kern-0.8em\TeX}}}

\copyrightyear{2021}
\acmYear{2021}
\setcopyright{acmlicensed}\acmConference[DIS '21]{Designing Interactive Systems Conference 2021}{June 28-July 2, 2021}{Virtual Event, USA}
\acmBooktitle{Designing Interactive Systems Conference 2021 (DIS '21), June 28-July 2, 2021, Virtual Event, USA}
\acmPrice{15.00}
\acmDOI{10.1145/3461778.3462078}
\acmISBN{978-1-4503-8476-6/21/06}

\begin{document}

\title{Giving Space to Your Message:\\Assistive Word Segmentation for the Electronic Typing of Digital Minorities}

\author{Won Ik Cho}
\affiliation{%
	\institution{Dept. of ECE and INMC, Seoul National University}
	\streetaddress{1 Gwanak-Ro}
	\city{Seoul}
	\country{Republic of Korea}
}
\email{wicho@hi.snu.ac.kr}
\author{Sung Jun Cheon}
\affiliation{%
	\institution{Dept. of ECE and INMC, Seoul National University}
	\streetaddress{1 Gwanak-Ro}
	\city{Seoul}
	\country{Republic of Korea}
}
\email{sjcheon@hi.snu.ac.kr}
\author{Woo Hyun Kang}
\affiliation{%
	\institution{Dept. of ECE and INMC, Seoul National University}
	\streetaddress{1 Gwanak-Ro}
	\city{Seoul}
	\country{Republic of Korea}
}
\email{whkang@hi.snu.ac.kr}
\author{Ji Won Kim}
\authornote{This research was done when the author was in Seoul National University.}
\affiliation{%
	\institution{IMS, University of Stuttgart}
	\country{Germany}
}
\email{st176776@stud.uni-stuttgart.de}
\author{Nam Soo Kim}
\affiliation{%
	\institution{Dept. of ECE and INMC, Seoul National University}
	\streetaddress{1 Gwanak-Ro}
	\city{Seoul}
	\country{Republic of Korea}
}
\email{nkim@snu.ac.kr}
\renewcommand{\shortauthors}{Cho et al.}

\begin{abstract}
  For readability and disambiguation of the written text, appropriate word segmentation is recommended for documentation, and it also holds for the digitized texts. If the language is agglutinative while far from scriptio continua, for instance in the Korean language, the problem becomes more significant. However, some device users these days find it challenging to communicate via key stroking, not only for handicap but also for being unskilled. In this study, we propose a real-time assistive technology that utilizes an automatic word segmentation, designed for digital minorities who are not familiar with electronic typing. We propose a data-driven system trained upon a spoken Korean language corpus with various non-canonical expressions and dialects, guaranteeing the comprehension of contextual information. Through quantitative and qualitative comparison with other text processing toolkits, we show the reliability of the proposed system and its fit with colloquial and non-normalized texts, which fulfills the aim of supportive technology. 
\end{abstract}





\maketitle

\section{Introduction}

Digitized text is an essential medium of communication nowadays, and compared to handwriting, readability is more likely to be obtained by various preprocessing such as enlarging and double spacing the letters and lines. Nonetheless, word segmentation is still a cumbersome issue since it regards the word morphology, which may affect the syntax and semantics of the text. Thus, the study on word segmentation\footnote{Although \textit{spacing} is more appropriate for Korean, we use \textit{segmentation} instead for consistency.} has been done for many languages including Hindi \cite{garg2010segmentation}, Arabic \cite{lee2003language}, Chinese \cite{chen2015gated}, Japanese \cite{sassano2002empirical}, and Korean \cite{lee2002automatic}. 

Although readability and machine comprehensibility matter, the necessity for segmentation depends on the language. For instance, in scriptio continua such as Japanese, segmentation is not obligatory in the writing process. However, in contemporary Seoul Korean, the spaces between words\footnote{The unit for spacing in Korean is called \textit{eojeol} (어절). It corresponds with \textit{word} in English.} are indispensable for the readability of the text \cite{lee2013automatic} and possibly for the naturalness of the speech synthesized thereof.

\begin{figure}
	\centering
	\includegraphics[width=0.6\columnwidth]{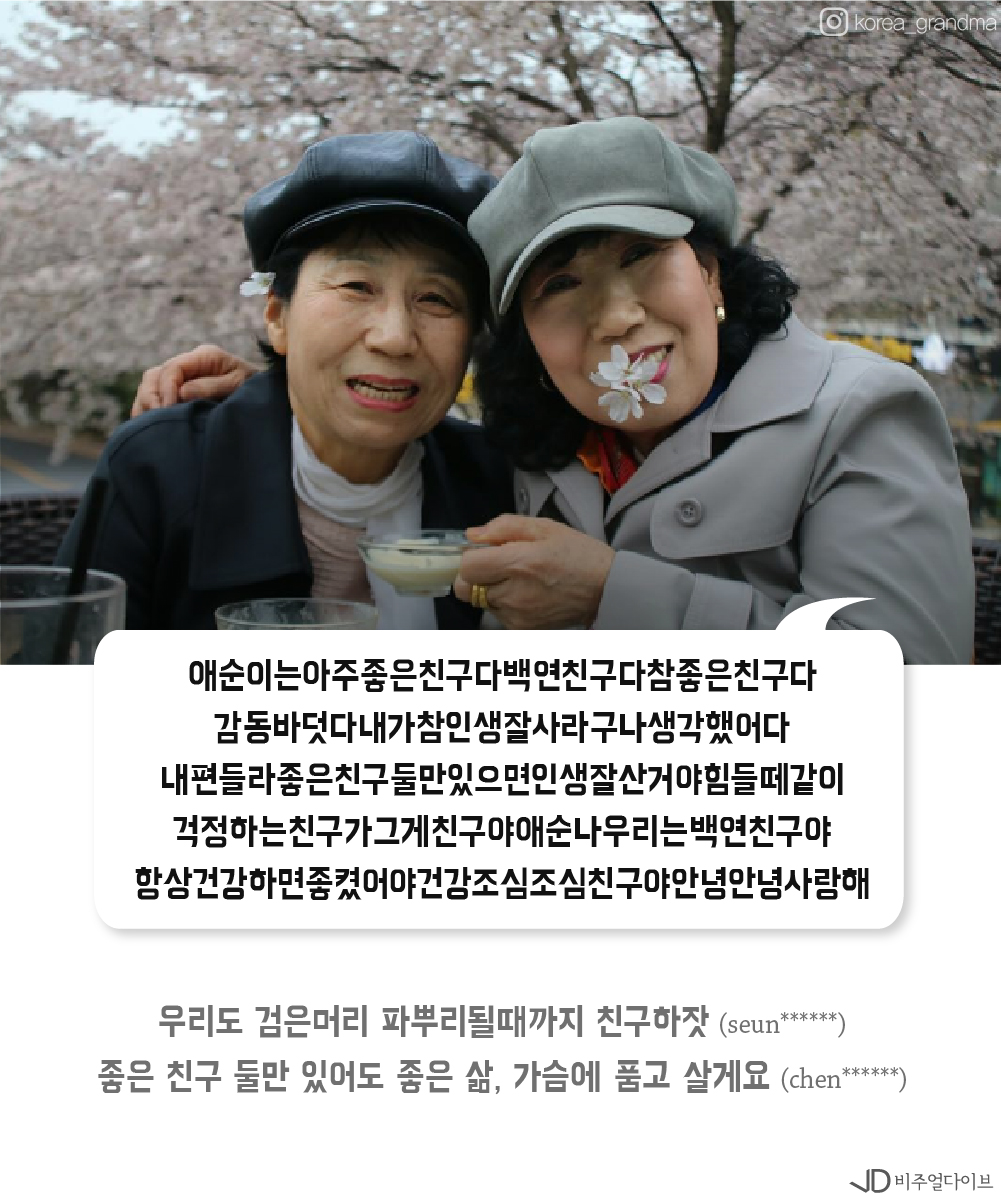}
	\caption{Instagram of an elderly Korean Youtube star, Makrae Park (Image by \textit{Visualdive}; \url{http://www.visualdive.com/}). Fluent utilization of mechanical segmentation system can be difficult for some digital minorities so that the resulting document lacks readability.} \label{fig:fig2}
\end{figure} 

We discerned that, in current language life where input through a device is universal, giving space with additional typing can be a time-consuming process. For example, in real-time chatting via desktop or mobile keyboards, some users enter non-segmented texts which lack readability and sometimes cause ambiguity. It prevents readers from concentrating on the text and makes it difficult for further natural language understanding (NLU) modules to perform the desired action. This often happens to originate from carelessness. However, we also observed that sometimes this unintentionally takes place, as a challenging issue for digital minorities such as the disabled or the elderly (Fig. 1), especially when they try to get familiar with the up-to-date electronic devices such as mobile phone. We were inspired by the technical needs that the readability issue should be resolved with an assistive, automatic word segmentation that preserves genuine user intention.

In this paper, we propose an assistive technology that aids device users in assigning proper spaces for non-segmented documents. The system is distinguished from conventional correction tools that evaluate the suitability of the input text concerning the language rules and suggest a grammatically appropriate one. From a slightly different perspective, the proposed system assigns proper spaces between the characters of a non-segmented input utterance in a non-invasive way that avoids distortion in the user’s intention. Although the resulting segmentation may not be strictly correct since ignoring the segmentation rules is often tolerated for informal conversation-style sentences in Korean, we expect that our system can enhance readability and provide naturalness in the viewpoint of prosodic break and speech production \cite{jung2007prediction}.

Our study encompasses the following as distinguishing features:

\begin{itemize}
	\item A real-world problem-based motivation which owes to the suffers of the digital minorities that come from the cumbersome characteristics of the Korean script typing
	\item Corpus construction, model architecture, and web application methodologies appropriate for real-time automatic segmentation of the noisy user-generated text
	\item A comparative study with the conventional spacing toolkits; a proposal of a subjective measure and a rigorous qualitative analysis utilizing sample scripts
\end{itemize}

\section{Background and Motivation}

Two main branches of our research regard assistive technology and natural language processing (NLP). Since the assistive point of view on word segmentation has rarely been discussed, here we first focus on the technologies that enable the system to comprehend the sentence-level contextual information.

Many studies on automatic segmentation of Korean text have dealt with statistics-based approaches on the previous words, correctly predicting the space that follows each character\footnote{Korean writing system \textit{Hangul} consists of about 2,500 characters, as to be demonstrated in Section~\ref{hangul}.}  \cite{lee2002automatic,lee2014balanced}. This issue is different from the studies on morphological analysis \cite{lee2011three}, though some approaches utilized decomposition into morphemes \cite{lee2002automatic}. In a recent implementation of \textit{KoSpacing} \cite{heewon2018}, convolutional neural network (CNN) \cite{kim2014convolutional} and bidirectional gated recurrent unit (BiGRU) \cite{chen2015long} were employed, utilizing various news articles in training. 

For Japanese, which shares a similar syntax with Korean, traditional approaches to the segmentation of the text and Kanji  sequence adopted statistical models \cite{sassano2002empirical,ando2003mostly}. However, contextual information does not seem to be fully utilized so far. In contrast, many contextual approaches have been suggested for Chinese word segmentation, although the embedding methodology and syntax are different from Korean. In \citet{chen2015gated}, context information was aggregated via a gated reculsive neural network to predict a space between characters. In \citet{cai2016neural}, a gated combination neural network was used over characters to produce distributed representations of word candidates.

In this way, the segmentation technique for non-spaced documents evolved briskly in the languages with indecisive spacings such as Chinese, Japanese, and Korean (CJK). However, we observed that most systems concentrate on increasing the appropriateness of the output, in terms of character-level or word-level accuracy, while less on how it can enhance the user experience, affect the following NLU systems, and finally facilitate the society with a data-driven artificial intelligence (AI). We considered that such approaches should incorporate the diversity of user expressions (beyond formal expressions), to whom and to what extent the service should be provided, and consideration of the domain in which it would be supportive.

The assistive point of view is considered in some NLP services provided along with products. The representative one in Korean is probably \textit{Naver Smart Keyboard}\footnote{\url{https://apps.apple.com/ai/app/naver-smartboard/id1297816298}}, which supports auto-completion in typing the frequent inputs. However, its main purpose is lessening the tiresomeness of the mechanical typing, rather than processing the unexpected inputs coming from immature control. \textit{Google Docs}\footnote{\url{https://docs.google.com/}} also has a similar auto-completion, but it acts more like grammatical or lexical error correction, rather than like segmenting the words to convey the meaning.

\section{Our Approach}

In this section, referring to the previous discussions, we investigate the adequate architecture that generates acceptable segmentation of user-generated texts, even when the users are not familiar with electronic typing. Our approach regards mainly two distinct features: (1) a conversation-style, linguistically diverse, and non-normalized corpus used for training, and (2) a concise and robust user-friendly automatic segmentation module.

\subsection{Corpus}
\label{corpus}

We use a training corpus distinguished from the widely used formal, written documents such as Wikipedia. Our corpus consists of dialogs in Korean drama scripts, semi-automatically collected, with punctuations and narration removed. It contains 2M utterances of various lengths and involves a wide variety of topics. The number of appearing characters is around 2,500, incorporating almost all the Korean characters used in real life. Several characteristics differentiate our dataset from the conventional corpora.
\paragraph{Linguistic diversity} The drama script is highly conversation-style, and a significant portion of dialogs incorporates expressions that do not fit with written Korean. This may be beneficial to the users who are not familiar with typing the standardized form of written language. It also includes various Korean dialects, middle-age Korean, and transliteration of non-Korean languages (e.g., Japanese and Chinese).
\paragraph{Non-normalizedness} A significant portion of the sentences contain non-normalized expressions. The corpus includes murmur, new words, slangs, incomplete sentences, etc., which enable the model to learn the expressions that are not covered in standard written web texts.
\paragraph{Weak segmentation rules}  Word (\textit{eojeol}) segmentation does not strictly follow the standard Korean as in literary language, but the corpus was human checked with the spaces that are inappropriately inserted. The omission of the spacing was tolerated in a manner that does not induce awkwardness regarding the prosody of the sentence if spoken. At least two Korean natives checked all the segmentation.
\paragraph{Allows typo} Typos were not corrected if the pronunciation does not affect the semantics of the sentence; this can be a strong point in the word segmentation of noisy user-generated texts, which are not usually checked for the errors. These include the spelling errors coming from mobile phone typing (e.g., \textit{too} and \textit{to} in English) or conventionally used non-standard lexical form (e.g., \textit{though} and \textit{tho} in English).\medskip

All the raw texts were manually searched and downloaded from publicly available websites where the users share the drama scripts. Though we cannot redistribute it due to the license issues, using it as a training corpus does not threaten the policies on intellectual property, given that our model does not generate nor replicate the materials.

\begin{figure*}
	\centering
	\includegraphics[width=\textwidth]{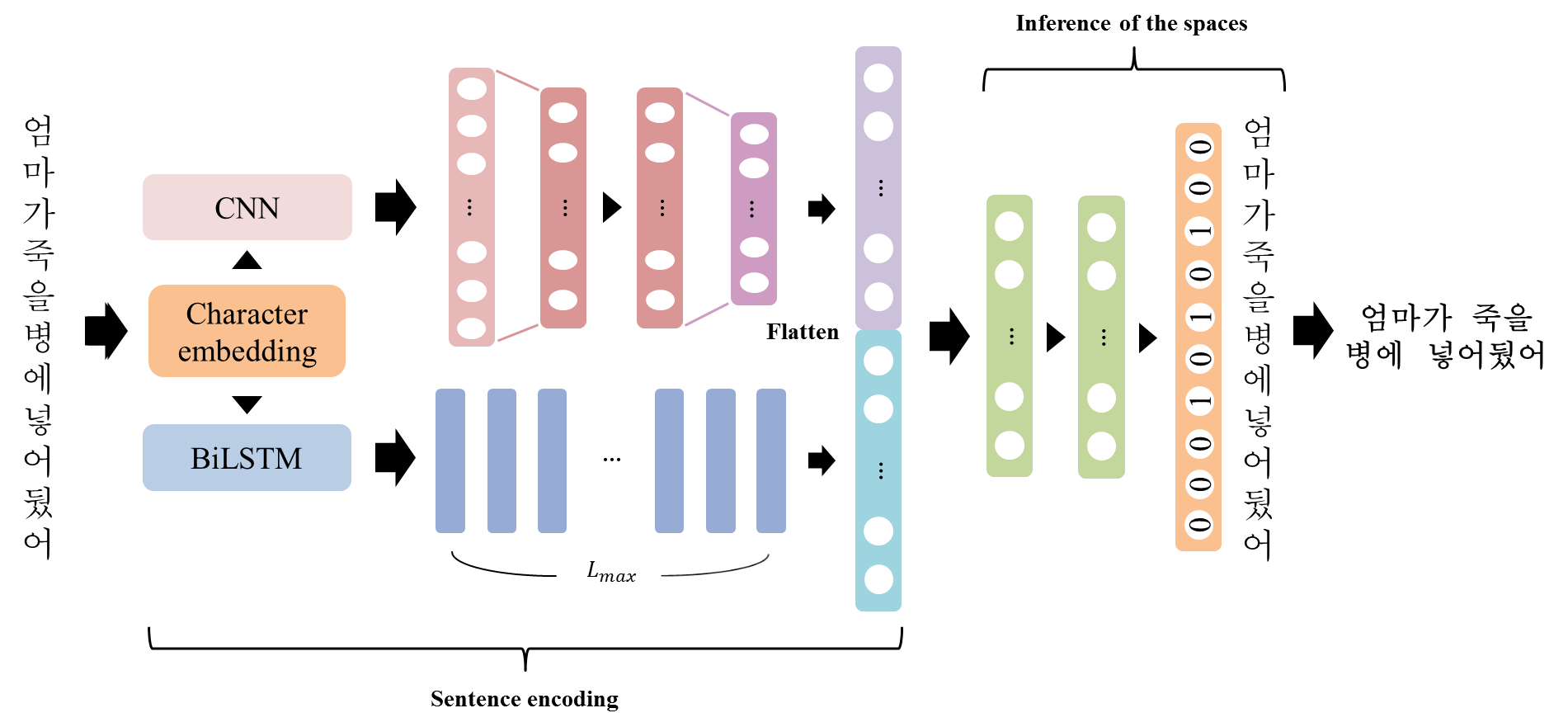}
	\caption{The architecture of the automatic segmentation module. The input sentence means \textit{``Mother put soup into the bottle''}, but if it were not for the space flag in the third place of the inference module, the sentence could have meant \textit{``I put the skin of mother into the bottle''}, which is  syntactically possible output for an ambiguous sentence but awkward. \textit{$L_{max}$} denotes the maximum character length of the input.} \label{fig:fig2}
\end{figure*} 

\subsection{Automatic segmentation module}

The next step is to train a system that automatically segments the input text. The segmentation module which is inspired by the encoder-decoder network \cite{cho2014learning} consists of two submodules, respectively managing \textit{sentence encoding} and \textit{space inference}.

\subsubsection{Character-based sentence encoding}
\label{hangul}

To clarify the writing system of the Korean language, we introduce the Korean character \textit{Hangul} and its components (alphabet) \textit{Jamo}. Roughly, \textit{Ja} denotes the consonants and \textit{mo} denotes the vowels, making up a morpho-syllabic block of \textit{Hangul} by \{Syllable: CV(C)\}. Note that the blocks are different from morphemes, which can be whatever form of a combination of \textit{Jamo}. We choose the character-level approach since the tokenization to characters is always identical, unlike the morpheme-level decomposition which is affected by the toolkit used. It does not request any tokenizer other than reading the Unicode string. This primarily demonstrates the conciseness and robustness of our approach.

For architecture, we use the concatenated network of CNN and Bidirectional long short-term memory (BiLSTM) \cite{hochreiter1997long} since they each represent the distributional and sequential summarization regarding the given input sentence (Fig. 2). We do not adopt the advanced NLP structures such as BERT \cite{devlin2019bert} here since the segmentation task requires less world knowledge, and it is assumed that such pre-trained knowledge may deter the advantage of using our corpora which does not overlap with the widely used sources such as Wikipedia. 

On this basis, the vector embedding of the characters is done based on skip-gram \cite{mikolov2013distributed}, an NLP technique that embeds the words into numerical representation concerning the relationship between the words, implemented with the subword scheme of fastText \cite{bojanowski2017enriching}. The pretraining was conducted on the corpus constructed in Section~\ref{corpus} as well\footnote{The word vector dictionary is be distributed publicly, also found in \url{https://github.com/warnikchow/raws}.}. Sparsely embedded vectors such as one-hot encoding are not exploited since they seemed inefficient for either CNN or BiLSTM, in the sense that they lack distributional information and usually require higher computation. Note that the number of possible characters constructed in the Korean language reaches 10K owing to the unique featural writing system of \textit{Hangul}. Also, the context-related properties of distributional semantics are expected to compensate for user-generated text noise, such as errata.

As a detailed procedure, the sentence is fed as an input to CNN and BiLSTM in the form of a non-segmented sequence of character vectors. They yield a flattened vector as an output layer of each network, which are concatenated to make up the encoding of the whole input sentence (Fig. 2). Roughly, this conveys our study’s concept as a contextual segmentation that can help disambiguate challenging composites. The system splits a sentence into words considering the whole composition, both the word history and the neighborhood. This enables the system to deal with the sentences containing ambiguous expressions, some to be discussed in Section~\ref{examples}. We expect CNN and BiLSTM can ensure efficient computation and play an essential role in this process. Their usage, placement, and hyperparameters were determined experimentally, and we checked that the proposed structure outperforms the cases of using either CNN or BiLSTM only, in terms of the prediction accuracy with our corpus. The specification on the model architecture is stated in Appendix~\ref{architecture}.

\subsubsection{Decoder with a binary output}

The backend of the segmentation module includes a decoder network that assigns proper segmentation for a given sequence of character vectors. With the vectorized encoding of the sentence as an input, the decoder network infers a binary sequence that indicates if there should be a space that follows each character. Note that the length of the output sequence equals $L-1$ where $L$ denotes the number of characters in the input sentence (See Appendix~\ref{architecture}).

\section{Experiment}

\subsection{Implementation}

Implementation for the system was done with  fastText library\footnote{\url{https://pypi.org/project/fasttext/}} and Keras \cite{chollet2015keras}, which were used for embedding character vectors and making neural network models, respectively. With GTX Tesla M40 24GB, the training for both systems took about 30 hours, two hours per epoch.

\subsection{Evaluation}

For evaluation, two widely-used text processing tools, namely (i) a correction module by \textit{Naver}\footnote{\url{https://bit.ly/3aGxPBr}} and (ii) \textit{KoSpacing}\footnote{Demo and model architecture are available in \url{https://github.com/haven-jeon/PyKoSpacing}}\cite{heewon2018} were adopted. \textit{Naver} does not disclose the architecture that is utilized, but it seems to base on a hybrid structure of rules and statistical models using Korean word taxonomy, considering the system malfunction induced by probably unseen expressions (e.g., dialect and slangs). \textit{KoSpacing} bases on a deep hierarchical structure of CNN and BiGRU but adopts character n-gram and various news articles. We treat each module as a representative of a word-based and character-based toolkit.

\subsubsection{Evaluation by preference}

For a subjective measure that suffices the purpose of this study, we noted the indecisiveness of the Korean word segmentation, which might fit with the proposed system that does not aim strictly accurate segmentation. In this regard, to scrutinize the user preference, we modified the concept of mean opinion score (MOS) to evaluate the result, which incorporates multiple answers.

For the survey, 30 posts were excerpted from the Instagram posts of Makrae Park\footnote{\url{https://www.instagram.com/korea_grandma}} (Figure 1, licensed). The test posts are not necessarily single utterances; they contain compound sentences which incorporate scrambling, dialects, non-normalized expressions, and many typos. Since the word-based correction modules did not respond with many of such samples, we considerately chose the posts with which the modules generate a segmented output. For the cases where the characters of the output were \textit{modified} due to the function of the module, the result was revised so that the characters in the original text are preserved. 

Provided with three sentences as the output of each system, namely \textit{Naver}, \textit{KoSpacing}, and \textit{Proposed}, 14 Korean natives were asked to assign the rank considering the naturalness of the segmentation. In the scoring phase, assigning the same rank to the outputs of more than one system was allowed. For example, (1,1,3) denotes two 1\textsuperscript{st}-places and one 3\textsuperscript{rd}-place. Similarly, (1,2,2) denotes two 2\textsuperscript{nd}-places. Also, to lessen the ambiguity, we only provided the samples where the output for all
the three systems differ from each other\footnote{The survey sheet is available in \url{https://github.com/warnikchow/raws/blob/master/survey.xlsx}.}.

For each sample $s_n$, for each system output $o_m$, we define the sample-wise system ranking as:

\begin{equation}
rank_{s_n,o_m} = avg(rank_p)
\end{equation}

the average rank for all the participant answer $rank_p$, with the participant pool $P = 14$. Next, for each sample, we set a system performance $SP$ which is defined by:

\begin{equation}
SP_{s_n,m} = \frac{M+1-rank_{s_n,o_m}}{M}
\end{equation}

for $rank$ the sample-wise preference and the number of the systems $M = 3$. Afterward, the final performance $SP\textsuperscript{*}$ of each system is yielded by averaging the system performance for the whole input utterances. That is, for each system, we have:

\begin{equation}
SP\textsuperscript{*}_{m} = \frac{\sum_n{SP_{s_n,m}}}{N} 
\end{equation}

for $N = 30$ the total number of the samples. The system rank was designed to result in 1.0 if everyone gives first place to a specific candidate and $\frac{1}{M}$ for the opposite case.

\subsubsection{Evaluation by accuracy}

Though the previous Instagram corpus displays how the proposed model can help maintain the naturalness of segmentation in challenging inputs, the evaluation cannot be done quantitatively since there exists no ground truth. Thus, notwithstanding the different training settings, corpus, and the aim of all the candidate modules, we deemed it beneficial to display the evaluation conducted with a fixed test set that contains conversational sentences. For this purpose, a chat log was collected in-house and removed with the special symbols. The total size is 3,127, with all spaced sentences in their raw form. The detailed procedure on the collection process is organized as a full experiment in a separate article \cite{seo2020thesis}.

Two types of utterance set are prepared; namely \textit{Easy} and \textit{Hard}. Here, in \textit{Easy}, the 50\% of the spaces in the original utterance are randomly removed, and in \textit{Hard}, all the spaces are removed. The evaluation is done on the F1 score, which tells whether the binary prediction per each character is correct or not; correct if the following slot is appropriately filled or blank concerning the original sentence. Especially for the cases where the input characters are modified, namely in \textit{Naver}, the evaluation only takes a look at the results of segmentation.

\begin{table}[]
	\centering
	\caption{The quantitative evaluation result.}
	\resizebox{0.45\columnwidth}{!}{%
		\begin{tabular}{|c|c|c|c|}
			\hline
			\textbf{System}             & \textit{\textbf{Naver}} & \textit{\textbf{KoSpacing}} & \textit{\textbf{Proposed}} \\ \hline
			\textit{\textbf{Easy (F1)}} & $\underline{0.9544}$            & 0.9186                      & 0.9090                     \\ \hline
			\textit{\textbf{Hard (F1)}} & $\underline{0.9333}$            & 0.8800                      & 0.9090                     \\ \hline
			\textit{\textbf{SP*}}       & 0.6809                  & 0.7142                      & $\underline{0.7444}$               \\ \hline
		\end{tabular}%
	}
	\label{tab:result}
\end{table}

\subsubsection{Results}

The results show that \textit{Naver} displays better scores compared to the deep learning and character-based models in accuracy-based evaluation (Table~\ref{tab:result}), though there had been some modification in the morphology. This originates from that conventional statistics-based and token-wise model well segments the lexicons. However, in view of content preservation, we cannot guarantee that the genuine intention of the input utterance is conveyed appropriately. This may harm the user satisfaction with mechanical modification.

Among \textit{KoSpacing} and \textit{Proposed}, which are two character-based approaches, \textit{KoSpacing} performs better in \textit{Easy} while not in \textit{Hard} (Table~\ref{tab:result}). Note that the F1 is not prone to be fluctuated by the text quality in the proposed model due to the distinguished training scheme. We deem this robustness is an advantage of the proposed model, which is less influenced by the errata or diversity of the user input.

The results of the subjectivity test are also proposed in Table~\ref{tab:result}. We interpret the results as, despite not being the most accurate system of all, the proposed system shows the highest naturalness over the others. We want to claim that the result does not imply the superiority of the proposed system towards the standard toolkits regarding the segmentation performance. Instead, it is suggested that the proposed system yields quite tolerable segmentation for the input utterances that are non-normalized and non-standard, and this is expected to be helpful for those who are not accustomed to electronic typing.

\subsubsection{Qualitative analysis}
\label{examples}

For qualitative analysis, We introduce the cases that display the pros and cons of using the proposed system. Here, for each sentence, the ground truth (GT) text is provided with proper segmentation, Yale romanization, gloss\footnote{HON honorific, PST past tense, DEC declarative, POL politeness, NOM nominative, and NMN nominalizer.}, and translation. Then, the input sentence is exhibited without segmentation, followed by the output from the proposed and another module. Note that the meaning of the output sentence equals that of the GT text unless stated otherwise.

\enumsentence{ \shortex{10}
	{나 [na] & 너 [ne] & 본 [pon] & 지 [ci] & 한 [han] & 세 [sey] & 달 [tal] & 다 [ta] & 돼 [tway] & 감 [kam]}
	{I & you & meet & from & about & three & months & almost & has & been}
	{`\textit{It's been almost three months since we last met.}'}
	\toplabel{sago}
	\item \shortex{1}
	{- \textit{Input}: 나너본지한세달다돼감}
	{- \textit{Naver}: 나 / 너 / 본 / 지 / 한 / 세 / 달 / 다 / 돼 / 감}
	{- \textit{Proposed}: 나 / 너 / 본지 / 한 / 세달 / 다 / 돼 / 감}
}
For (1), \textit{Naver} shows the result  that adheres to the
segmentation rules. In contrast, \textit{Proposed} yields an output in which some characters comprising the phrases are agglutinated. Although strictly correct output is grammatically appropriate, the output from the proposed model aggregates the components that are syntactically close (e.g., \textit{pon-ci} (meet-from, since the last met), \textit{sey-tal} (three-months)), making the prosodic durations more natural (in view of pronunciation). This seems to be the result influenced by the weak segmentation rules of the training script, which yields the corpora that the other models utilize might lack.%\medskip\\

\enumsentence{ \shortex{10}
	{아버지 [apeci] & 친구분 [chinkwu-pwun] & 당선되셨어요 [tangsen-toy-sy.ess-e-yo]}
	{father & friend-HON & elect-be-HON.PST-DEC-POL}
	{`\textit{Father's friend was elected.}' (honorific, polite)}
	\toplabel{sago}
	\item \shortex{1}
	{- \textit{Input}: 아버지친구분당선되셨어요}
	{- \textit{KoSpacing}: 아버지 / 친구 / 분당선 / 되셨어요 (Father's friend became Bundang line.)}
	{- \textit{Proposed}: 아버지 / 친구분 / 당선 / 되셨어요}
}
Sentence (2) incorporates morphological and semantic ambiguity; if a space goes in between `구 (\textit{kwu})' and `분 (\textit{pwun})', not between `분 (\textit{pwun})' and `당 (\textit{tang})', the sentence meaning changes to ``\textit{Father's friend became Bundang line\footnote{A subway line of Korea.}.}'' which is syntactically acceptable but semantically awkward. In the input, \textit{KoSpacing} inserts a space between \textit{kwu}/\textit{pwun} and \textit{Proposed} puts between \textit{pwun}/\textit{tang}; in the case of \textit{KoSpacing}, where \textit{chinkwu} (friend) becomes apart with its honorific suffix \textit{pwun} and \textit{pwun} makes up a new noun phrase \textit{pwuntangsen} (Bundang line), completely different meaning is conveyed although the other part of the output sentence is the same as that of \textit{Proposed}. This kind of segmentation seems to happen because \textit{KoSpacing} was trained with the news articles which more embody the topics regarding transportation than colloquial expressions.%\medskip\\

\enumsentence{ \shortex{10}
	{뭣이 [mwes-i] & 중헌지도 [cwunghen-ci-to] & 모름서 [molum-se]}
	{what-NOM & important-NMN-also & ignorant-DEC}
	{`\textit{You don't even know what's important!}' (Southwestern Korean dialect)}
	\toplabel{sago}
	\item \shortex{1}
	{- \textit{Input}: 뭣이중헌지도모름서}
	{- \textit{Naver}: 뭣이중헌지도모름서}
	{- \textit{Proposed}: 뭣이 / 중헌지도 / 모름서}
}
For (3), \textit{Naver} does not make any change regarding the unseen data, Southwestern dialect which may not have been incorporated in the training corpus. \textit{Proposed} shows a sound output, without omission of a character nor a transformation of the sentence.

\begin{figure*}[h]
	\centering
	\includegraphics[width=\textwidth]{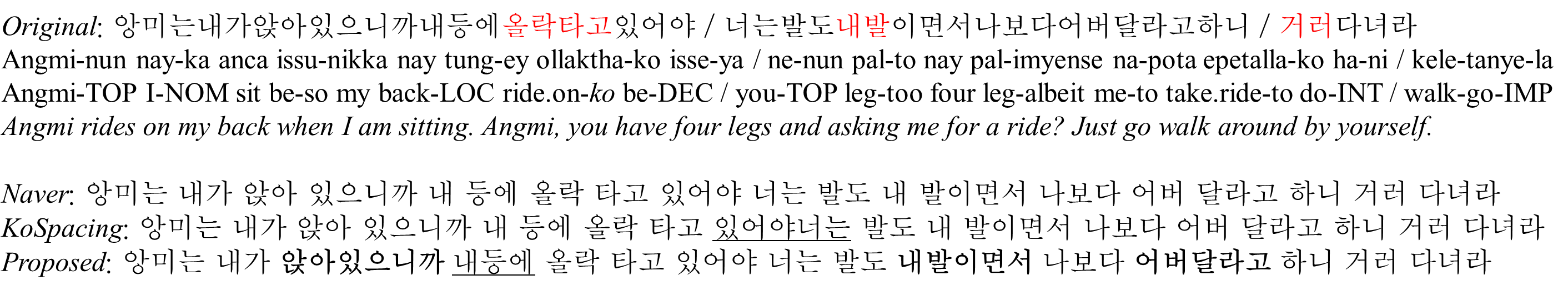}
	\caption{A sample post (\textit{Original}), its romanization, gloss, translation, and output sentences. Some typing errors of the user are colored in red in the original text but were not taken into account in the translation. \textit{Angmi} refers to a dog. The input for the systems is identical to \textit{Original}, except for the removal of slashes (`/') which were added to help the readers recognize the sentence boundary.} \label{fig:fig2}
\end{figure*}

As a more challenging case, we provide a sample post in the survey sheet (Fig. 3). Given a post text from the Instagram account, which incorporates non-normalized expressions, dialects, and errata, the output sentences from the baseline modules and the proposed system are listed at the bottom of the figure. The underlined phrases in \textit{KoSpacing} and \textit{Proposed} denote the parts that are not correct, displaying the demerit of the two systems concerning \textit{Naver} module in the sample. This kind of errors led the  \textit{Proposed} to score the second place in the given sentence\footnote{\textit{Naver} the first and \textit{KoSpacing} the third.}. However, this single case does not hinder our aim of assigning naturalness and readability to the noisy input utterances rather than aiming at an accurate correction. For instance, three bold chunks in  \textit{Proposed} denote the results that are distinct in the proposed model. It exhibits the naturalness regarding syntax and prosody, even given that the first two phrases are not strictly grammatical. Although not all the cases are displayed here, the proposed system exhibits the result that fits the goal of our implementation, especially with the user-generated inputs and adversarial examples.%, even with errata and some profanity terms.

%\enumsentence{ \shortex{10}
%	{앙미는 [angmi-nun] & 내가 [nay-ka] & 앉아 있으니까 [anca issu-nikka] & 내 등에 [nay tung-ey] & 올락타고 있어야 [ollaktha-ko isse-ya] & 너는 [ne-nun] & 발도 내발이면서}
%	{Angmi-TOP & I-NOM & sit be-so & my back-LOC & ride.on-\textit{ko} be-DEC & you-TOP }
%	{`\textit{You don't even know what's important!}' (Southwestern Korean dialect)}
%	\toplabel{sago}
%	\item \shortex{1}
%	{- \textit{Input}: 뭣이중헌지도모름서}
%	{- \textit{Naver}: 뭣이중헌지도모름서}
%	{- \textit{Proposed}: 뭣이 / 중헌지도 / 모름서}
%}

%During the survey, the participants were asked to check all the appropriate and incorrect parts of the output sentences. For instance, 

\section{Discussion}

So far, we have displayed how the technical side of our approach can be beneficial to our goal, making up an assistive word segmentation by yielding acceptable spaces that preserve the intent and enhance the readability. Some questions may arise again: How is this different from the grammatical error correction?
Will this truly benefit digital minorities? By what mean can this be applied in the real-world scenario? 

First, we want to note again that our approach differs from the \textit{correction} that can modify the users’ intention inadvertently. We focus on avoiding unwanted revision, which prevents the users from correcting the modified sentence once again to fit their will, as seen in Sentence (1) in Section~\ref{examples} where the grammatical sentence does not guarantee naturalness. This can be interpreted as a process of compensating grammaticality to the level of acceptability for the sake of naturalness.

%\begin{figure*}
%	\centering
%	\includegraphics[width=0.7\textwidth]{fig5}
%	\caption{Some search result images of `\textit{grandfather chat message}' in Korean, where the segmented text are typed owing to the digital non-friendliness of the elderly citizens.} \label{fig:fig2}
%\end{figure*} 

Next, we claim that our approach is not invasive, given that the target type of text is yet the ones with the lowest readability possible (as in Fig. 1). Since our system can transform a non-segmented text to at least a more segmented one, this also lessens the ambiguity of the document from the highest degree to somewhat moderate status. One concern is, there can indeed be the case that the author could have intended to make such a non-segmented document for some purpose, e.g., to raise ambiguity or to express one's feelings. It often happens among the users, and sometimes to utilize the given typing space as efficiently as possible. However, at least in our observation, most cases regarding the digital minorities seemed unintentional%\footnote{\url{https://www.picuki.com/media/2186021057387627571}}\footnote{\url{https://www.insight.co.kr/news/137213}} (Fig. 4)
. Though those typings may move the readers' heart by conveying the author's will at messaging, we inferred that there should be many more undisclosed cases where they had found the segmentation issues disturb their communication.

Lastly, the application of this system is not limited to assistive technology for digital minorities. For instance, in Korean, the proposed system receives any sentence input and generates the segmented one which fits with prosody, aside from the grammatically correct ones \cite{kim2007role}. Thus, automatic speech recognition (ASR) or text-to-speech (TTS) modules of intelligent devices may utilize the proposed function as a post- or pre-processing respectively, to provide more plausible output \cite{yoon2006prosodic}.

\section{Application}

Our system has the potential to be utilized as an assistive interface for various web-based services such as Instagram and Telegram. Although not directly related to the system’s performance, mainly three issues are argued for industrial potential.

\paragraph{Long sentences}

Since the overall architecture of the proposed system deals with the sentences of fixed length  \textit{$L_{max}$}, which is set to 100 (Appendix~\ref{architecture}), the inference of spaces for long sentences requires an additional procedure. To cover such cases, we apply an overlap-hopping for the sentences whose length exceeds \textit{$L_{max}$}. The overlap length is 30 (characters), and the decision of the segmentation follows the inference that is done later. For example, to simplify the description, if the input is given `A B C D E F G H I J '. where $L_{max}$ is 5, and overlap length is 1, the segmentation is performed in the order of `A B C D E', `B C D E F' $\cdots$ `F G H I J' and then concatenated. This kind of approach is expected to bring more satisfying result compared to merely splitting the whole sequence into `A B C D E' and `F G H I J' and performing two-fold segmentation.

\paragraph{Segmentation given by the user}

Though we aim to cope with the user input’s non-segmentation issue, we may also face cases that the user has given some spacings in the input already. Our trained module ignores such information in the inference process, as described in the previous section. However, we thought it beneficial to take the user’s desire into account, as in \citet{lee2014balanced}. If the user input contains the spaces between some characters, then it precedes the trained module’s inference. We implement this as an optional function, whose execution can be decided by the program supervisor or the user.

\begin{figure*}
	\centering
	\includegraphics[width=\textwidth]{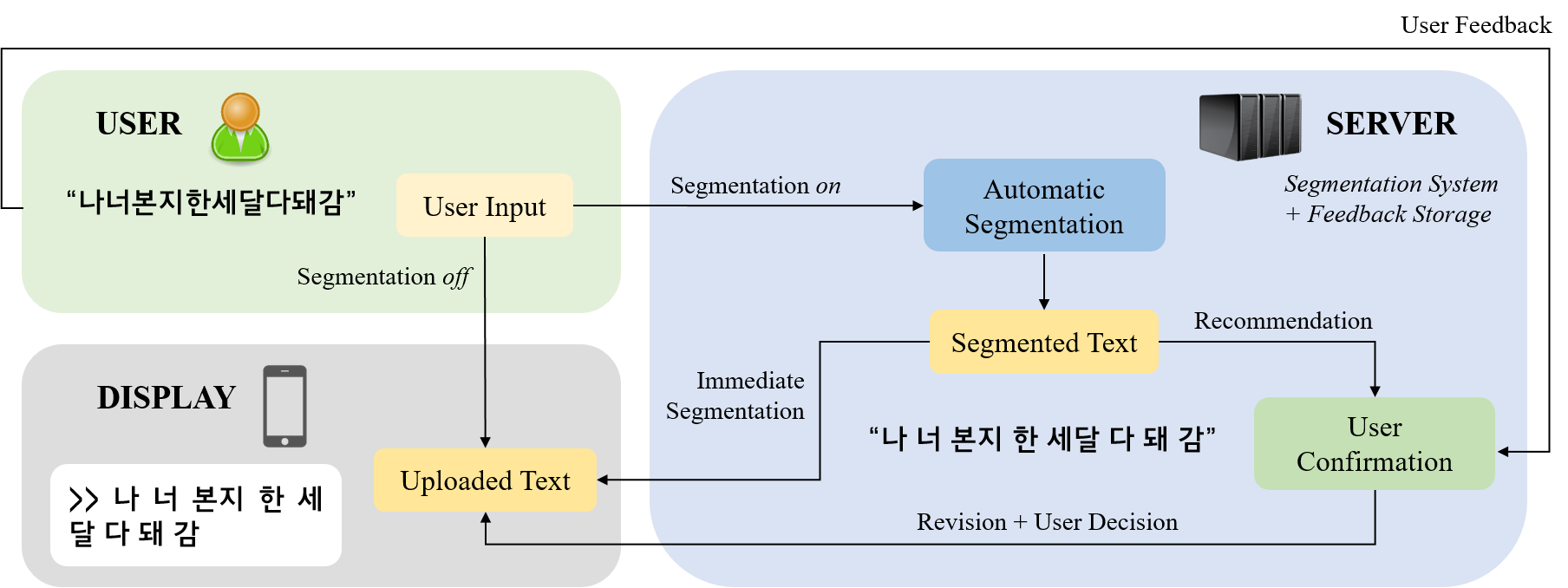}
	\caption{A brief sketch on the  web-based segmentation service. The sample sentence comes from the example (1) of Section~\ref{examples}.} \label{fig:fig2}
\end{figure*} 

\paragraph{Immediate segmentation and recommendation}

Our system incorporates a deep structure, so the issue of computation should not be overlooked. Thus, we suggest a system that can act as either an immediate segmenter or a segmentation recommender, depending on the option user selects (Fig. 4). Since a large portion of web-based services accompanies a procedure of restoring query on the server, a non-segmented text the user generates can be primarily fed as an input to the automatic segmentation module, which is running online. If the user wants an immediate segmentation, the module directly uploads a segmented version of the sentence to the display. If the user wants a recommendation, the segmented version is returned so that the user can make a revision and confirmation on it. This user-friendly function has a lot to do with the motivation of this study. It is not concretely implemented here but is expected to be easily materialized in the user interface.

\section{Conclusion}

In this paper, we suggested an assistive word segmentation system that can serve as a real-time text editor for noisy user-generated text. First, we refined Korean drama scripts, making up 2M lines respectively, in the way that the segmentation results fit with the prosodic and contextual naturalness of the text. Based on it, a data-driven architecture was trained to yield an automatic segmenter of which the performance was verified by quantitative and qualitative comparison with widely used text processing modules. For Korean, where the strict rule is not necessarily required to obey in real-time written conversation, we proposed a subjective measure that can compare the naturalness of the segmentation that the systems assign to user-generated texts. The results support our system's fit on the aim of assistive technology, also assuring the robustness in the objective measure. The discussion was performed on how our approach differs from grammatical correction, whether this technology really matters, and with which features it can be utilized in the real world. Adopting the overlap-hopping method and leveraging the user segmentation information may prevent the text processing system from dismissing the user's will.

In future work, the whole context of the user-generated document can be engaged in to deal with the segmentation that incorporates more noise and further ambiguity. We also encourage this scheme for other writing systems that may face a similar issue in the typing of digital minorities. The pre-trained system is available online\footnote{\url{https://github.com/warnikchow/raws}}, and the fastText character vector dictionary trained with the corpus is also distributed for the research purpose\footnote{Disclosing the corpus is limited due to the copyright.}.

\begin{acks}

This research was supported by the Technology Innovation Program (10076583, Development of free-running speech recognition technologies for embedded robot system) funded By the Ministry of Trade, Industry \& Energy (MOTIE, Korea). The authors appreciate Makrae Park, Visualdive, ASAP of Hanyang University, and Ye Seul Jung for providing the motivation, infographics, corpus, and useful example sentences. Also, the experimental results would not have been available without the grateful contribution of Suin Seo, along with the ChatSpace project.

\end{acks}

\newpage
\bibliographystyle{ACM-Reference-Format}
\bibliography{dis2021}

\newpage
\appendix

\section{Network Architecture}
\label{architecture}

\begin{figure*}[h]
	\centering
	\includegraphics[width=0.85\textwidth]{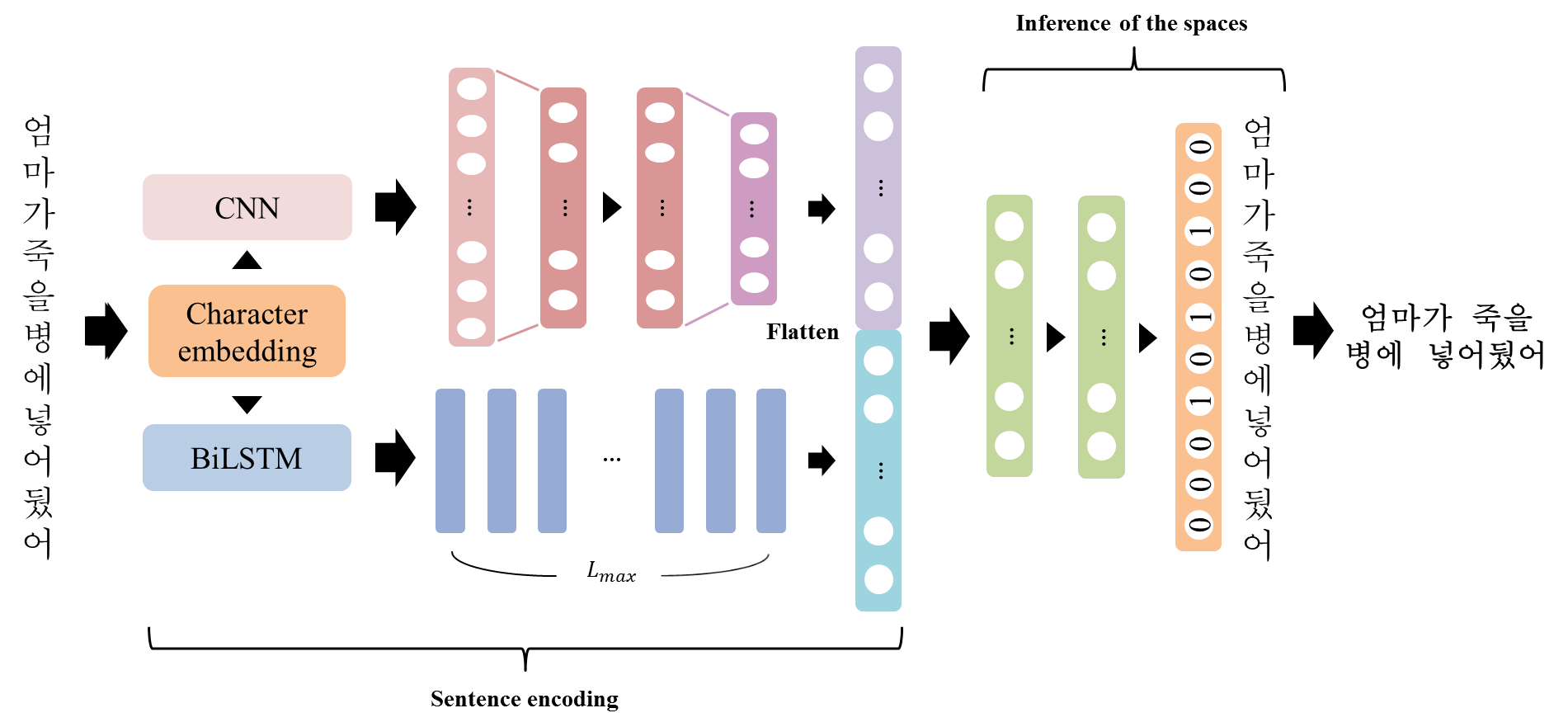}
	\caption*{Fig. 2} \label{fig:fig2}
\end{figure*} 

In the architecture, \textit{$L_{max}$} considers the usual length of the utterances. The input is the feature extracted from a non-segmented utterance, and the output is a vector with binary entries that indicate the predicted segmentation. Note that the activation and loss function (ReLU, MSE) were chosen experimentally. For Korean, using sigmoid resulted in overfitting that caused an awkward segmentation for even a simple composite. We assume this phenomenon comes from the volume of the character set (\textgreater 2500). The threshold regarding MSE was set to 0.5.

\begin{table}[h]
	\caption{System specification.}
	\resizebox{0.5\columnwidth}{!}{%
		\begin{tabular}{|c|c|c|}
			\hline
			& \multicolumn{2}{c|}{\textbf{Specification}} \\ \hline
			\multirow{2}{*}{\textbf{Corpus}} & \multirow{2}{*}{\begin{tabular}[c]{@{}c@{}}Instance of \\ utterances\end{tabular}} & \multirow{2}{*}{2,000,000 (\#char: \textgreater{}2,500)} \\
			&  &  \\ \hline
			\multirow{4}{*}{\textbf{CNN}} & \begin{tabular}[c]{@{}c@{}}Input size\\ (single channel)\end{tabular} & ($L_{max}$, 100, 1) \\ \cline{2-3} 
			& Filters & 32 \\ \cline{2-3} 
			& Window & \begin{tabular}[c]{@{}c@{}}Conv layer: (3, 100)\\ Max pooling: (2,1)\end{tabular} \\ \cline{2-3} 
			& \# Conv layer & 2 \\ \hline
			\multirow{3}{*}{\textbf{BiLSTM}} & \multirow{2}{*}{Input size} & \multirow{2}{*}{($L_{max}$, 100)} \\
			&  &  \\ \cline{2-3} 
			& Hidden layer nodes & 32 \\ \hline
			\multirow{2}{*}{\textbf{MLP}} & Hidden layer nodes & 128 \\ \cline{2-3} 
			& \# Layers & 2 \\ \hline
			\multirow{6}{*}{\textbf{Others}} & Optimizer & Adam (0.0005) \\ \cline{2-3} 
			& Batch size & 128 \\ \cline{2-3} 
			& Dropout & 0.3 (for MLPs) \\ \cline{2-3} 
			& Activation & ReLU (MLPs, CNN, output) \\ \cline{2-3} 
			& \multirow{2}{*}{Loss function} & \multirow{2}{*}{Mean squared error (MSE)} \\
			&  &  \\ \hline
		\end{tabular}%
	}
	\label{my-label}
\end{table}

\end{document}